# 1.5 billion words Arabic Corpus

## Ibrahim Abu El-khair

*Information Science Dept., Faculty of Social Sciences, Umm Al-Qura University-KSA*
*LIS Dept., Faculty of Arts, Minia University-Egypt*
*iabuelkhair@gmail.com*

## Abstract

This study is an attempt to build a contemporary linguistic corpus for Arabic language. The corpus produced, is a text corpus includes more than five million newspaper articles. It contains over a billion and a half words in total, out of which, there is about three million unique words. The data were collected from newspaper articles in ten major news sources from eight Arabic countries, over a period of fourteen years. The corpus was encoded with two types of encoding, namely: UTF-8, and Windows CP-1256. Also it was marked with two mark-up languages, namely: SGML, and XML.

## 1 Introduction

The efficiency of any information retrieval systems mainly depends on the experiments conducted by the researchers in the field, and commercial companies producing these systems. These experiments are done to emulate real world queries submitted to any system and the response of it to these queries. It is usually conducted in a closed laboratory environment. Elements of the retrieval process in this type of experiments are controlled by the researchers, in order to determine causes of success or failure and fixing it.

Language corpora are one of the most important elements for information retrieval experiments in particular and for natural language processing in general. This is because the corpus represents the actual everyday use of the language. Corpus use in retrieval has improved significantly in most languages especially Latin based languages. As for Arabic language it is still relatively new.

Arabic Language is the language of the holy Quran. It is used by more than a billion and a half Muslims around the world in the daily rituals. It is the mother tongue of about two hundred and fifty million people around the world. It is also, the official language of twenty-two countries and an official language for non-Arabic countries like Chad, Eretria, Mali, and Turkey (Encyclopaedia Britannica Almanac, 2009). Moreover, it is one of the six official languages of the United Nation (UN, 2015), since 1973 (UN, 1973).

In spite of all of the above, Arabic language Corpora still in need for more research and studies. There is an ongoing need for more Arabic Corpora. The majority of available corpora now are relatively small in size, or rather expensive. The main purpose of this paper is producing a new free corpus. A corpus with a large size, representative of the language, from different countries, different writing styles, from more than one source, and distributed over many years. It will be available for researchers in the field of information retrieval, computational linguistics, and natural language processing.

## 2 Available Arabic Corpora:

Table one shows some of the previous attempts to create Arabic corpora. It should be noted that the review will be limited to textual monolingual corpora, not word lists, lexicons, speech, and opinion corpora, all types were reviewed by Zaghouani, (2014).

## 3 Data Collection:

Web scrapping or web copying programs were used to extract text from news sources in order to create the corpus. The researchers used wget [1], which is used by LDC, and htttrack [2] site copier, but both were very slow, so they were not used. Two other program, Internet Download Manager [3], cyotek webcopy [4], were used and eliminated as well because they stop working for no apparent reason, in addition to being slow. After several attempts the researcher used MetaProducts Offline Explorer Pro[5], Visual Web Ripper [6]. Both programs were very good in extracting text and eliminating all unnecessary objects like images, videos, JavaScript files, and CSS files.

### 3.1 Corpus Sources:

There are a lot of news sources that could be used for creating a language corpus. At this paper, the researcher has chosen ten sources to be used in the corpus. Several news websites were tested before selecting the source that will be used. The fame of the website, and the news source, or the number of readers were not the criterion for selection. There were other criteria and technical reasons for selecting the news resources used in building the corpus.

---

[1]. https://www.gnu.org/software/wget
[2]. https://www.httrack.com
[3]. https://www.internetdownloadmanager.com
[4]. http://www.cyotek.com/cyotek-webcopy
[5]. http://www.metaproducts.com/mp/offline_explorer_pro.htm
[6]. http://www.visualwebripper.com

- The first criterion is having no overlap with previous Arabic corpora. For example, Al-Ahram newspaper from Egypt has the largest digital news archive on the internet, but were not selected because it is a part of the Arabic Gigaword Corpus.
- The source should be online for a long time. This is simply to have a large volume

| No. | Corpus | Words | Texts | Unique Words | Licensing | Data Type |
|---|---|---|---|---|---|---|
| 1 | Current Corpus | 1525722252 | 5222973 | 3303723 | Free | Newspaper articles |
| 2 | Arabic Gigaword, 5$^{th}$ ed., (26) | 1077382000 | 3346167 | Unavailable | $ 6000 | Newspaper articles |
| 3 | Arabic Gigaword, 4$^{th}$ ed., (25) | Unavailable | 2716995 | 848 469 | $ 5000 | Newspaper articles |
| 4 | Arabic Gigaword, 3$^{rd}$ ed., (16) | Unavailable | 1994735 | 576 799 | $ 4000 | Newspaper articles |
| 5 | Arabic Gigaword, 2$^{nd}$ ed., (17) | Unavailable | 1591987 | 481 906 | $ 3000 | Newspaper articles |
| 6 | Arabic Gigaword, 1$^{st}$ ed., (15) | Unavailable | 1256719 | 391 619 | $ 3000 | Newspaper articles |
| 7 | King Abdulaziz City for Science and Technology (KACST) Corpus (11) | 732780509 | 869 800 | 7464396 | Free | Multiple |
| 8 | An-Nahar Newspaper Text Corpus (12) | 144 million | 270000 | Unavailable | € 504 | Newspaper articles |
| 9 | Arabic Modern Standard Corpus (3) | 113 million | 102 134 | Unavailable | Free | Newspaper articles |
| 10 | The International Corpus of Arabic (ICA) (7) | 79569384 | 70,022 | 1272766 | Free | Newspaper articles, books, emails |
| 11 | LDC Corpus (Arabic Newswire: part 1), (18) | 76 million | 383 872 | 666 094 | $ 1200 | Newspaper articles |
| 12 | King Saud University Corpus of Classical Arabic (KSUCCA) (9) | 50 million | Unavailable | Unavailable | Free | books Classic |
| 13 | Open Source Arabic Corpus (OSAC), (27) | 22 million | 32,262 | Unavailable | Free | Multiple |
| 14 | Al-Hayat Arabic Corpus, (8) | 18639264 | 42,591 | Unavailable | € 720 | Newspaper articles |
| 15 | Akhbar El-Khaleeg 2., (2, 14) | 10 million | Unavailable | Unavailable | Free | Newspaper articles |
| 16 | University of Jordan Arabic Corpus (UJAC), (19) | 7522941 | 61,037 | 707 385 | Free | Newspaper articles |
| 17 | Akhbar El-Khaleeg 1., (1) | 3 million | Unavailable | Unavailable | Free | Newspaper articles |
| 18 | Contemporary Arabic Corpus, (10) | 842 684 | 416 file | Unavailable | Free | Newspaper articles, websites' emails |
| 19 | NEMLAR Corpus, (24) | 500000 | Unavailable | Unavailable | € 300 | Multiple |
| 20 | Al-Raya Corpus, (4,5,6,20) | 219 978 | 187 | 30,096 | Free | Newspaper articles |
| 21 | SACS Corpus (Saudi Arabian National Computer Science Conference), (4,5,6,21) | 46,968 | 242 | Unavailable | Free | Research Abstracts |
| 22 | Arabic Corpus Project, (28,29) | Unavailable | 400 | Unavailable | Free | Books |

Table 1. Available Arabic Corpora

| Source (English) | Source (Arabic) | Abbrev. | Country | From | To | Website |
|---|---|---|---|---|---|---|
| Alittihad | الاتحاد الإماراتية | ETD | Emirates | Jan. 2008 | June 2014 | http://www.alittihad.ae |
| Echorouk Online | الشروق أون لاين | SHG | Algeria | Feb. 2008 | May 2014 | http://www.echo-roukonline.com/ara |
| Alriyadh | الرياض | RYD | KSA | Oct. 2000 | Dec. 2013 | http://www.alriyadh.com |
| Alyaum | اليوم | YMS | KSA | July 2002 | Dec. 2013 | http://www.alyaum.com |
| Tishreen | تشرين | TRN | Syria | Jan. 2004 | May 2014 | http://www.tishreen.news.sy |
| Alqabas | القبس | QBS | Kuwait | Jan. 2006 | April 2014 | http://www.alqabas.com.kw |
| Almustaqbal | المستقبل | MTL | Lebanon | Sep. 2003 | April 2014 | http://www.almustaqbal.com |
| Almasry-alyoum | المصري اليوم | MSY | Egypt | Dec. 2005 | Jan. 2014 | http://www.almasry-alyoum.com |
| youm7 | اليوم السابع | YM7 | Egypt | Jan. 2008 | May 2013 | http://www.youm7.com |
| Saba News Agency | وكالة أنباء سبأ اليمنية | SBN | Yemen | Dec. 2009 | May 2014 | http://www.sabanews.net |

Table 2. Corpus resources

| Source | Articles | | Words | | Unique Words | |
|---|---|---|---|---|---|---|
| | Number | Percentage | Number | Percentage | Number | Percentage |
| *Alriyadh* | 858,188 | 16.43% | 271,353,697 | 17.79% | 1,451,320 | 15.39% |
| *youm7* | 1,025,027 | 19.63% | 261,700,304 | 17.15% | 1,020,444 | 10.82% |
| *Alyaum* | 888,068 | 17.00% | 237,914,494 | 15.59% | 1,319,996 | 13.99% |
| *Alqabas* | 817,274 | 15.65% | 233,741,575 | 15.32% | 1,260,511 | 13.36% |
| *Alittihad* | 349,342 | 6.69% | 139,962,699 | 9.17% | 932,628 | 9.89% |
| *Almustaqbal* | 446,873 | 8.56% | 135,446,906 | 8.88% | 982,765 | 10.42% |
| *Tishreen* | 314,597 | 6.02% | 94,695,378 | 6.21% | 905,169 | 9.60% |
| *Almasryalyoum* | 291,723 | 5.59% | 93,398,135 | 6.12% | 760,511 | 8.06% |
| *Echorouk Online* | 139,732 | 2.68% | 40,978,911 | 2.69% | 543,799 | 5.77% |
| *Saba News Agency* | 92,149 | 1.76% | 16,530,153 | 1.08% | 255,098 | 2.70% |
| *Totals* | 5222973 | 100.00% | 1,525,722,252 | 100.00% | 3,303,723 | |

Table 3. Corpus Statistics according to the source.

of articles available. This was perhaps one the major obstacles in conducting this study. Knowing when the newspaper appeared online, was a problem. There was no way of knowing that without checking each one individually since there is no website that could have this information.
- All selected sources should represent different countries in the Arab world.
- The scrapped text should be in an editable form.
- The selected news source website should allow the crawling programs to work on it and import the articles. Some websites have very tight security procedures, and do not allow spidering.

It should be noted that the news websites crawling was done between December 2013 and June 2014. Two of the sites, almustaqbal, sabanews, were re-crawled because of errors discovered in the quality control phase. There was a problem importing the publication date in them.

Table two, indicates the selected sources for the corpus, its name in English and in Arabic, its abbreviation, the time period for each one of them, country of origin, and its website. Nine newspapers, and one news agency from eight countries were selected as shown in the table. Egypt and Saudi Arabia are represented with two newspapers each, since they are the pioneers in online journalism, and have some of the oldest online newspapers in the Arab world.

The coverage period varies from one source to the other. The starting time in each news source is basically the time it first appeared online. The ending date depended on the time of the data collection. Some websites allowed harvesting the news archive but not the current news like Alyaum from Saudi Arabia, and Almasryalyoum from Egypt.

### 3.2 Metadata:

Two tagging schemes were used with the corpus in hand. All articles in the current corpus were tagged with SGML (Standard Generalized Markup Language), which is used in TREC corpora. The other scheme was using XML (Extensible Markup Language) tagging, which is used in the LDC corpora.

Each article will have an ID using the source abbreviation, table one, Arabic language abbreviation, and a serial number, e.g.

<ID> RYD_ARB_0000001 </ID>, or

<DOCNO> RYD_ARB_0000001 </DOCNO>.

### 3.3 Encoding:

The corpus will be encoded with windows cp-1256[7] for Arabic language. It will also be encoded with UTF-8[8]. Having two versions of the corpus with two different encoding schemes will be of great use for researchers in the field of Arabic information retrieval, and Natural language processing.

## 4 Results

As mentioned earlier, the corpus by itself is useless unless it is used to serve some a research area. The main purpose for creating this corpus, is to have a free tool for Arabic language available for researcher. It is made specifically for work in the field of information retrieval, or natural language processing.

The corpus is not limited to one subject. It is multitopic news corpus covering Politics, literature, arts, technology, sports, economy, culture, and many other subject matters. It is also, a good representation of Arabic language. It covers a period of fourteen years and eight countries. These countries have a very large portion of Arabic native speakers. Finally, all ten sources used in creating the corpus are well represented.

Table three shows the statistics of the corpus in details, and what has been assembled from each source of ten sources. It includes the number and percentage of articles that have been imported from each source, and the total number and percentage of words and unique words for each source. It has been arranged based on the number of words; because they determine the value of each source for corpus. It should be noted that the total number of "unique words" is not equal to the addition of the values in the column; because all repeated words between sources are excluded.

## 5 Conclusion

Language corpus is a representation of the language use. It should be, according to Mansour's principles (2013), large, have a specific purpose, diverse, representative, and well balanced. In order to have a general idea about the corpus in hand, in terms of size. Table four, shows the general statistics of the corpus. It indicates that the corpus has over five million articles from ten news sources. The total number of words exceeds 1.5 billion words, and the total number of unique words exceeds 3.3 million words.

---

[7]. https://msdn.microsoft.com/en-us/goglobal/cc305149.aspx

[8]. http://unicode.org/resources/utf8.html

| Number of resources | Nine Newspapers, One news Agency |
|---|---|
| Number of countries covered | Eight Countries |
| Years covered | 14 Years |
| Corpus Size | 10GB (CP-1256 ) / 16GB (UTF-8) |
| Number of articles | 5,222,973 Articles |
| Number of Words | 1,525,722,252 Words |
| Number of Unique Words | 3,303,723 Words |

Table 4. General Statistics of the corpus

The KACST Corpus (Al-Thubaity, 2014), the largest free corpus available, created by a team from King Abdulaziz City for Science and Technology. They also outsourced 25% of the corpus to external specialists. It has 700 million words with about 1.5 million articles. The Arabic Giga-Word corpus, which is the largest paid corpus available, was created by an institution like the LDC over a period of over of ten years. It has 3.3 million articles, and 1.077 billion words.